\documentclass[conference]{IEEEtran}
\IEEEoverridecommandlockouts
\usepackage{cite}
\usepackage{amsmath,amssymb,amsfonts}
\usepackage{algorithmic}
\usepackage{graphicx}
\usepackage{textcomp}
\usepackage{xcolor}
\usepackage{multirow}
\usepackage{graphicx}
\usepackage{hyperref}

\def\BibTeX{{\rm B\kern-.05em{\sc i\kern-.025em b}\kern-.08em
    T\kern-.1667em\lower.7ex\hbox{E}\kern-.125emX}}
\begin{document}

\title{Dynamic Mode Decomposition based feature for Image Classification}

\author{\IEEEauthorblockN{Rahul-Vigneswaran K\textsuperscript{*}\textsuperscript{\dag} \thanks{\textsuperscript{*}Website : \href{https://rahulvigneswaran.github.io}{rahulvigneswaran.github.io}} \thanks{ \textsuperscript{\dag}Work done while interning at CEN}}
\IEEEauthorblockA{\textit{Department of Mechanical Engineering} \\
\textit{Amrita Vishwa Vidyapeetham}\\
Amritapuri, India \\
rahulvigneswaran@gmail.com}
\and
\IEEEauthorblockN{Sachin-Kumar S, Neethu Mohan, Soman KP}
\IEEEauthorblockA{\textit{Center for Computational Engineering and Networking (CEN)} \\
\textit{Amrita School of Engineering, Coimbatore}\\
Amrita Vishwa Vidyapeetham, India  \\
\{sachinnme, neethumohan.ndkm, kpsoman2000\}@gmail.com}

}

\maketitle

\begin{abstract}
Irrespective of the fact that Machine learning has produced groundbreaking results, it demands an enormous amount of data in order to perform so. Even though data production has been in its all-time high, almost all the data is unlabelled, hence making them unsuitable for training the algorithms. This paper proposes a novel method\footnote{All codes and datasets used in this paper are available at :\\ \href{https://github.com/rahulvigneswaran/Dynamic-Mode-Decomposition-based-feature-for-Image-Classification}{https://github.com/rahulvigneswaran/Dynamic-Mode-Decomposition-based-feature-for-Image-Classification}} of extracting the features using Dynamic Mode Decomposition (DMD). The experiment is performed using data samples from Imagenet. The learning is done using SVM-linear, SVM-RBF, Random Kitchen Sink approach (RKS). The results have shown that DMD features with RKS give competing results. 
\end{abstract} 


\section{Introduction}\label{introduction}
The human race is generating more data than any point of time in history. Still, they are unsuitable for training an algorithm because almost all the data generated are either biased or unlabelled. Due to these reasons, data scientists are being restricted to use only a minuscule portion of the generated data, that has been preprocessed and cleaned. That makes the almost 99\% of the generated data unusable and on the other hand, today's state-of-the-art Deep learning algorithms are designed to be data thirsty in its training stage. Recent days, scientist have started designing architectures that can learn the distribution with less data, unlike their counterparts. Largely, these new waves of algorithms can achieve this by the following ways,
\begin{itemize}
    \item Semi-supervised Learning \cite{1,2,3,4,5,6,7,8,9,10,11,12,13}
    \item Weak supervised Learning \cite{14,15,16,17,18,19,20}
    \item Active Learning \cite{21,22,23}
    \item Transfer Learning \cite{24,25}
    \item Multi-task Learning \cite{26,27}
    \item Few-shot Learning \cite{28,29,30}
    \item Data Augmentation \cite{31,32}
    \item Reinforcement Learning \cite{33,34}
\end{itemize}

\cite{1,3} comes under the category of generative models of data. As the name suggests, \cite{1} make use of the generative a classifier through a generative model by iterative Expectation-Maximization (EM) techniques, a variant of Deterministic Annealing whole and \cite{3} use the unlabelled data to make the synthetically generated labelled data, less synthetic by use of Generative Adversarial Nets (GANs)\cite{4}. \cite{5,6} uses a method called co-training (When a set of data is divided into parts by nature and this trait is exploited by algorithms, they are categorized into co-training) where \cite{5} finds the weak indicator from labelled data and finds the corresponding unlabeled data to strengthen it. Like the methods discussed so far, there are several techniques used for learning with limited labelled data. Table \ref{tab1} gives a detailed summary of methods from each category mentioned previously.

Section \ref{introduction} gives a brief introduction on the existing methods for limited labelled data learning, Section \ref{materials} provides an elaborate explanation of the concepts used in the proposed approach like Dynamic Mode Decomposition (DMD) and Random Kitchen Sink (RKS) algorithm. Section \ref{approach} details the proposed approach and Section \ref{results} elaborates the obtained results and draws the underlying commonalities which are interesting. Finally Section \ref{conclusion} gives an essence of the proposed approach's findings and concludes with the future scope of this research.

\begin{table*}[ht]
\caption{Summary of papers and their corresponding techniques which aid them in learning with limited labelled data.}
\centering
\resizebox{\textwidth}{!}{%
\begin{tabular}{|c|c|c|c|c|}
\hline
\textbf{Category} & \textbf{Sub-Category} & \textbf{Specific Method} & \textbf{Motivation} & \textbf{Paper} \\ \hline
\multirow{8}{*}{Semi-Supervised} & \multirow{4}{*}{Generative Models of Data} & \multirow{2}{*}{-} & Limited Labelled + Unlabelled & \cite{1} \\ \cline{4-5} 
 &  &  & Synthetic-Labelled + Real-Unlabelled & \cite{3} \\ \cline{3-5} 
 &  & \multirow{2}{*}{Co-Training} & \multirow{2}{*}{Limited Labelled + Unlabelled} & \multirow{2}{*}{\cite{5,6}} \\
 &  &  &  &  \\ \cline{2-5} 
 & Low-Density Separation & Transductive Learning & Labelling the Unlabelled using Labelled & \cite{8} \\ \cline{2-5} 
 & \multirow{3}{*}{Graph-Based} & \multirow{2}{*}{Label Propagation} & \multirow{2}{*}{Labelling the Unlabelled using Labelled} & \multirow{2}{*}{\cite{9,10}} \\
 &  &  &  &  \\ \cline{3-5} 
 &  & - & Completely Unlabelled & \cite{13} \\ \hline
\multirow{7}{*}{Weak Supervision} & Noisy Labels & Relation Extraction & Heuristic labelling of Completely Unlabelled data & \cite{14} \\ \cline{2-5} 
 & \multirow{4}{*}{Generative Models of Labels} & Relation Extraction & Removing Wrong labels from the Heuristic labelling of Completely Unlabelled data & \cite{15} \\ \cline{3-5} 
 &  & \multirow{3}{*}{-} & Limited Labelled + Large Weakly Labelled & \cite{16} \\ \cline{4-5} 
 &  &  & Labelling of Unlabelled data & \cite{17} \\ \cline{4-5} 
 &  &  & Error reduction of Labelled data & \cite{18} \\ \cline{2-5} 
 & Biased Labels & PU-Learning & Positive and Unlabelled data & \cite{19} \\ \cline{2-5} 
 & Feature Annotation & NA & Use Labelled features & \cite{20} \\ \hline
\multirow{3}{*}{Active Learning} & \multirow{2}{*}{-} & \multirow{2}{*}{-} & Human labels the required unlabelled data & \cite{21} \\ \cline{4-5} 
 &  &  & Transfering dataset & \cite{22} \\ \cline{2-5} 
 & - & Inductive Learning & Transfering model & \cite{25} \\ \hline
\multirow{2}{*}{Multi-Task Learning} & \multirow{2}{*}{-} & \multirow{2}{*}{Inductive Learning} & \multirow{2}{*}{Limited Labelled data} & \multirow{2}{*}{\cite{26,27}} \\
 &  &  &  &  \\ \hline
\multirow{2}{*}{Few-shot Learning} & \multirow{2}{*}{-} & \multirow{2}{*}{-} & \multirow{2}{*}{Limited Labelled data} & \multirow{2}{*}{\cite{28,29}} \\
 &  &  &  &  \\ \hline
\multirow{2}{*}{Data Augmentation} & \multirow{2}{*}{-} & \multirow{2}{*}{-} & \multirow{2}{*}{Increase the Labelled data count} & \multirow{2}{*}{\cite{31,32}} \\
 &  &  &  &  \\ \hline
\multirow{2}{*}{Reinforcement Learning} & - & Apprenticeship Learning & Learning directly from the Expert without the need for any dataset & \cite{33} \\ \cline{2-5} 
 & - & Policy Shaping & Modifying policy in realtime by getting advice from a human & \cite{34} \\ \hline
\end{tabular}%
}

\label{tab1}
\end{table*}

\section{Materials and Methods}\label{materials}

\subsection{Dataset}\label{dataset}

The dataset used for benchmarking is the Tiny Imagenet Dataset which is a miniature version of the Imagenet Dataset. It contains 200 classes and each class contains 500 Images each. Each Image is 64x64 pixels in size.

\subsection{Dynamic Mode Decomposition (DMD)}\label{dmd}

It's a way of extracting the underlying dynamics of a given data that flows with time. It is a very powerful tool for analysing the dynamics of non-linear systems and was developed by Schmid \cite{35}. It is also used for 
forecasting \cite{49}, natural language processing \cite{50}, salient region detection from images \cite{38}, etc. It was inspired by and closely related to Koopman-operator analysis \cite{36}. The popularity gained by DMD in the fluids community is majorly due to its ability to provides information about the dynamics of flow, even when those dynamics are inherently non-linear. In short, DMD is a method driven by data, free from the equation which has the capability of providing a precise decomposition of a system which is highly complex into respective coherent spatio-temporal structures, that can be fashioned for predicting for few timestamps into the future. A typical DMD algorithm involves the following enumerations,

\begin{enumerate}
    \item Compute Singular Value Decomposition (SVD) of $X_a$ as,
    \[{X_a} \approx Q\Sigma {B^H}\]
    There $Q \in {C^{N \times J}}$, $\Sigma  \in {C^{J \times J}}$, $B \in {C^{M \times J}}$, J refers to the SVD approximation of $X_a$ which is reduced.
    
    \item Compute matrix C from,
    \[{X_a} \approx {X_b}C \Rightarrow Q\Sigma {B^H}C\]
    \[C \Rightarrow B{\Sigma ^\dag }{Q^H}{X_b}\]
    
    \item Compute similar matrix of $C$ which is ${\tilde C}$ by,
    \[\tilde C \Rightarrow {Q^H}{X_b}{\Sigma ^\dag }B\]
    
    \item Compute the Eigen Decomposition of $\tilde C$ by,
    \[AQ\Sigma {B^H} = {X_b} \Rightarrow AQ = {X_b}B{\Sigma ^\dag }\]
    Pre-multiply by ${Q^H}$ on both sides,
    \[{Q^H}AQ = {Q^H}{X_b}B{\Sigma ^\dag } = \tilde C\]
    \[AQ = Q\tilde C = Q\left( {T\Omega {T^\dag }} \right) \Rightarrow A\left( {QT} \right) = \left( {QT} \right)\Omega \]
    There, ${T\Omega {T^\dag }}$ is the eigen decomposition.
    
    \item Compute the Dynamic Modes matrix by,
    \[\Phi  = QT \Rightarrow \Phi  = {X_b}B{\Sigma ^\dag }T\]

\end{enumerate}

\subsection{Random Kitchen Sink algorithm}\label{rks}

The aim of Random Kitchen Sinks (RKS) algorithm and the methods similar to it, is not to perform inference but rather aim at overcoming the limitations of other kernel-based algorithms.

Kernel-based algorithms perform well in almost all the settings but heavily depend on matrix manipulation. If a matrix is $(n \times n)$ then naively the computation cost is $O({n^3})$ which bottlenecks them to applications that have limited samples. One of the general ways to overcome this limitation is by use of low-rank methods (even though other approaches like Bayesian committee machines and Kronecker based methods exist).

Random Fourier features \cite{37} aims at sampling subset components of kernel Fourier to generate low-rank-approximations of kernels that are invariant to shifts. Due to the reason that the Fourier spaces are invariant to shift, this property is not changed. But now a kernel Hilbert space which is reproduced by a finite dimensional space by these Fourier components' union. As a result, the RKS which was infinite dimensional once is approximated by the degenerate approximate kernel.

The epitome of supervised machine learning approaches are to obtain the knowledge of an approximate function $\tilde f (priori)$ which can map the input variable $\tilde x$ to output variable $\tilde y$ (i.e.) $\tilde y = \tilde f(\tilde x)$. The idea of finding such a function is that when a new data comes, the function can predict the corresponding output. In real-world applications, the input data can be image, 1-D signal, text data etc and the output will be the corresponding labels. The learning of mapping function often involves finding the best parameters $\theta $ for the function $\tilde f(\tilde x;\theta )$ to get the maximum performance. Kernel methods are the best examples of supervised
approaches which extensively used for several machine learning problems. It requires to compute a Kernel matrix ${R_{M \times M}}$ ($M$ signifies the count of the input vectors). However, the above mentioned computation suffers badly when the dataset size is large. There has been an effort to reduce the dimension of the Kernel matrix using
smart sample selection \cite{39}, Eigen decomposition via Nystrom \cite{40}, low-rank approximations \cite{41}. In \cite{42, 43}, the authors proposed an alternative approach via randomization, known as Random Kitchen Sinks (RKS) algorithm, to compute the Kernel matrix even when the dataset size becomes large. The idea is to provide an approximate kernel function via explicit mapping

\[K\left( {{x_a},{x_b}} \right) = \left( {\phi ({x_a}),\phi ({x_b})} \right) \approx \left( {z({x_a}),z({x_b})} \right)\]

Here, $\phi ( \bullet )$ denotes the implicit mapping function (used to compute kernel matrix) and $z( \bullet )$ denotes the explicit mapping function. The RKS method approximates the kernel trick \cite{44,45}. This explicit mapping function can be written as \cite{46,47,48}.

\[z(x) = \sqrt {\frac{1}{k}} \left[ {\begin{array}{*{20}{c}}
{\begin{array}{*{20}{c}}
{\cos \left( {{x^T}{\Omega _1}} \right)}\\
 \vdots 
\end{array}}\\
{\cos \left( {{x^T}{\Omega _k}} \right)}\\
{\sin \left( {{x^T}{\Omega _1}} \right)}\\
{\begin{array}{*{20}{c}}
 \vdots \\
{\sin \left( {{x^T}{\Omega _k}} \right)}
\end{array}}
\end{array}} \right]\]

\section{Proposed Approach}\label{approach}

\begin{figure}[htbp]
\centerline{\includegraphics[width=70mm,scale=0.8]{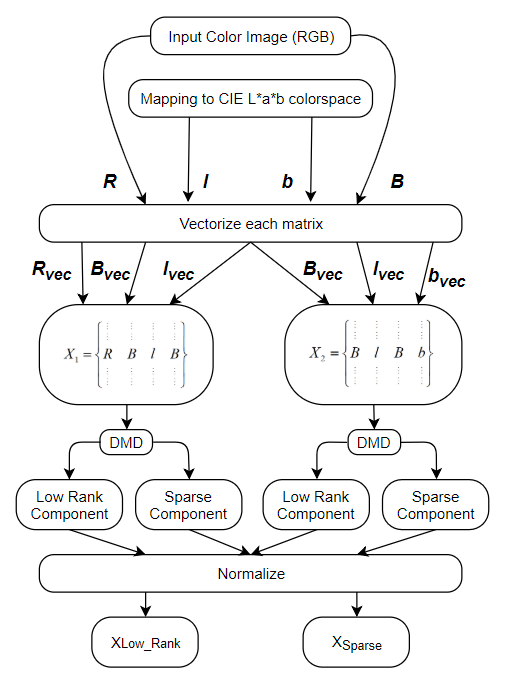}}
\caption{Proposed Architecture}
\label{fig1}
\end{figure}

As mentioned earlier in Section \ref{dmd}, DMD can be used in data that can flow in time. Contrary to that, the images that we are using are static in nature. Therefore flow is induced \cite{38} to the image as shown in Figure \ref{fig1}, by extracting the different bands of colours by converting it into Lab colour space and permutation of luminescence bands and colour bands into a single matrix. After applying DMD, the sparse and low-rank components are extracted and normalized for being used as features.

These extracted features capture the underlying dynamics of the image. Therefore, these are then given as input to the Random Kitchen Sink Algorithm (Section \ref{rks}) for classification.

\section{Results and Discussion}\label{results}
After obtaining the features, they are given as input to the Random Kitchen Sink algorithm and Support Vector Machine (SVM) with various configurations. Tables \ref{tab2}, \ref{tab3}, \ref{tab4} contain the accuracies of the proposed approach under various configurations when classified by Random Kitchen Sink algorithm, SVM-rbf and SVM-linear respectively. Column 1 represents the percentage of the total dataset used for setting, column 2 represents the count of eigenvalues taken into consideration for reconstruction of the features, column 3 denotes the type of class (Distinctive - Classes that are easily differentiable from one another; Overlapped - Classes that have overlapping features), and column 4 represents the accuracy of the corresponding configuration. All the accuracies corresponding to their configurations are plotted in Figure \ref{fig2}. It is evident from it that the Random Kitchen Sink Algorithm (RKS), constantly comes on top as compared to the other two algorithms. 

\begin{table}[]
\centering
\caption{Accuracy of Random Kitchen Sink Algorithm under different configurations}
\label{tab2}
\begin{tabular}{|c|c|c|c|}
\hline
\textbf{\begin{tabular}[c]{@{}c@{}}Test Data\\ (in \%)\end{tabular}} & \textbf{\begin{tabular}[c]{@{}c@{}}Number of \\ Eigen Values\end{tabular}} & \textbf{Type of Data} & \textbf{\begin{tabular}[c]{@{}c@{}}Accuracy\\ (in \%)\end{tabular}} \\ \hline
70 & 3 & Distinctive & 69.57 \\ \hline
70 & 3 & Overlapped & 57.50 \\ \hline
70 & 4 & Distinctive & 67.34 \\ \hline
70 & 4 & Overlapped & 67.55 \\ \hline
70 & 5 & Distinctive & 73.14 \\ \hline
70 & 5 & Overlapped & 61.31 \\ \hline
60 & 3 & Distinctive & 77.18 \\ \hline
60 & 3 & Overlapped & 53.47 \\ \hline
60 & 4 & Distinctive & 73.92 \\ \hline
60 & 4 & Overlapped & 60.69 \\ \hline
\textbf{60} & \textbf{5} & \textbf{Distinctive} & \textbf{80.87} \\ \hline
60 & 5 & Overlapped & 64.00 \\ \hline
50 & 3 & Distinctive & 72.41 \\ \hline
50 & 3 & Overlapped & 60.20 \\ \hline
50 & 4 & Distinctive & 76.85 \\ \hline
50 & 4 & Overlapped & 60.87 \\ \hline
50 & 5 & Distinctive & 80.52 \\ \hline
50 & 5 & Overlapped & 64.97 \\ \hline
\end{tabular}
\end{table}

\begin{table}[]
\centering
\caption{Accuracy of Support Vector Machine (rbf Kernel) under different configurations}
\label{tab3}
\begin{tabular}{|c|c|c|c|c|}
\hline
\textbf{\begin{tabular}[c]{@{}c@{}}Test Data\\ (in \%)\end{tabular}} & \textbf{\begin{tabular}[c]{@{}c@{}}Number of\\ Eigen Values\end{tabular}} & \textbf{Type of Data} & \textbf{Kernel} & \textbf{\begin{tabular}[c]{@{}c@{}}Accuracy\\ (in \%)\end{tabular}} \\ \hline
70 & 3 & Distinctive & rbf & 51.52 \\ \hline
70 & 3 & Overlapped & rbf & 34.76 \\ \hline
70 & 4 & Distinctive & rbf & 47.43 \\ \hline
70 & 4 & Overlapped & rbf & 34.00 \\ \hline
70 & 5 & Distinctive & rbf & 48.76 \\ \hline
70 & 5 & Overlapped & rbf & 36.86 \\ \hline
60 & 3 & Distinctive & rbf & 52.89 \\ \hline
60 & 3 & Overlapped & rbf & 34.89 \\ \hline
\textbf{60} & \textbf{4} & \textbf{Distinctive} & \textbf{rbf} & \textbf{55.00} \\ \hline
60 & 4 & Overlapped & rbf & 36.11 \\ \hline
60 & 5 & Distinctive & rbf & 50.56 \\ \hline
60 & 5 & Overlapped & rbf & 36.11 \\ \hline
50 & 3 & Distinctive & rbf & 52.40 \\ \hline
50 & 3 & Overlapped & rbf & 34.13 \\ \hline
50 & 4 & Distinctive & rbf & 51.33 \\ \hline
50 & 4 & Overlapped & rbf & 36.67 \\ \hline
50 & 5 & Distinctive & rbf & 54.00 \\ \hline
50 & 5 & Overlapped & rbf & 38.27 \\ \hline
\end{tabular}
\end{table}

\begin{table}[]
\centering
\caption{Accuracy of Support Vector Machine (Linear Kernel) under different configurations}
\label{tab4}
\begin{tabular}{|c|c|c|c|c|}
\hline
\textbf{\begin{tabular}[c]{@{}c@{}}Test Data\\ (in \%)\end{tabular}} & \textbf{\begin{tabular}[c]{@{}c@{}}Number of\\ Eigen Values\end{tabular}} & \textbf{Type of Data} & \textbf{Kernel} & \textbf{\begin{tabular}[c]{@{}c@{}}Accuracy\\ (in \%)\end{tabular}} \\ \hline
70 & 3 & Distinctive & linear & 46.38 \\ \hline
70 & 3 & Overlapped & linear & 34.00 \\ \hline
70 & 4 & Distinctive & linear & 42.10 \\ \hline
70 & 4 & Overlapped & linear & 32.86 \\ \hline
70 & 5 & Distinctive & linear & 42.95 \\ \hline
70 & 5 & Overlapped & linear & 36.10 \\ \hline
60 & 3 & Distinctive & linear & 47.78 \\ \hline
60 & 3 & Overlapped & linear & 34.00 \\ \hline
60 & 4 & Distinctive & linear & 44.33 \\ \hline
60 & 4 & Overlapped & linear & 34.22 \\ \hline
60 & 5 & Distinctive & linear & 42.44 \\ \hline
60 & 5 & Overlapped & linear & 34.44 \\ \hline
\textbf{50} & \textbf{3} & \textbf{Distinctive} & \textbf{linear} & \textbf{49.20} \\ \hline
50 & 3 & Overlapped & linear & 32.00 \\ \hline
50 & 4 & Distinctive & linear & 44.80 \\ \hline
50 & 4 & Overlapped & linear & 36.53 \\ \hline
50 & 5 & Distinctive & linear & 44.93 \\ \hline
50 & 5 & Overlapped & linear & 33.07 \\ \hline
\end{tabular}
\end{table}

It is evident from both the Figure \ref{fig2} and Table \ref{tab2} that, the maximum accuracy is attained when the configuration is as follows :\bigskip

\begin{tabular}{@{}lll}
   Test Data (in \%)      & : & 60\\
   Number of Eigen Values & : & 5\\
   Type of Data   & : & Distinctive
\end{tabular}\bigskip

\begin{figure}[htbp]
\centerline{\includegraphics[width=100mm,scale=1]{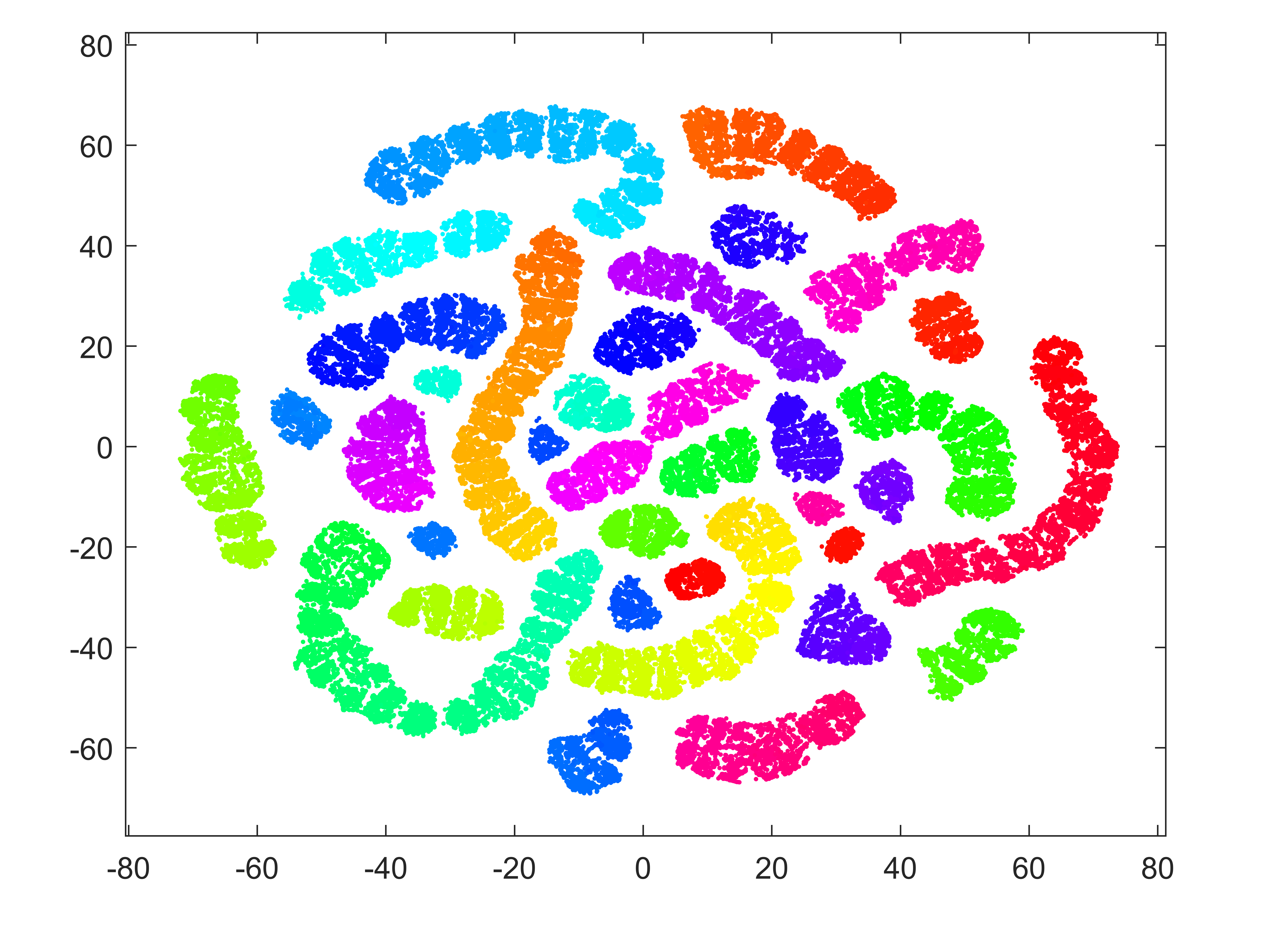}}
\caption{2-Dimensional t-SNE plot of the DMD feature of 50 randomly selected classes with considerations for 5 eigenvalues.}
\label{2dtsne}
\end{figure}

\begin{figure}[htbp]
\centerline{\includegraphics[width=100mm,scale=1]{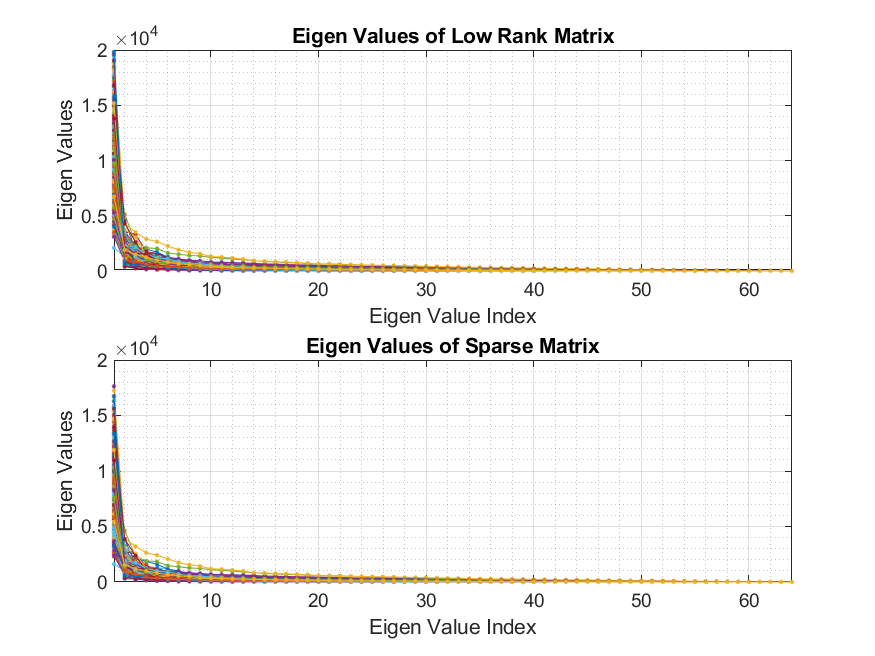}}
\caption{Eigen Value plot of Low rank and sparse matrix}
\label{fig3}
\end{figure}

\begin{figure}[htbp]
\centerline{\includegraphics[width=100mm,scale=1]{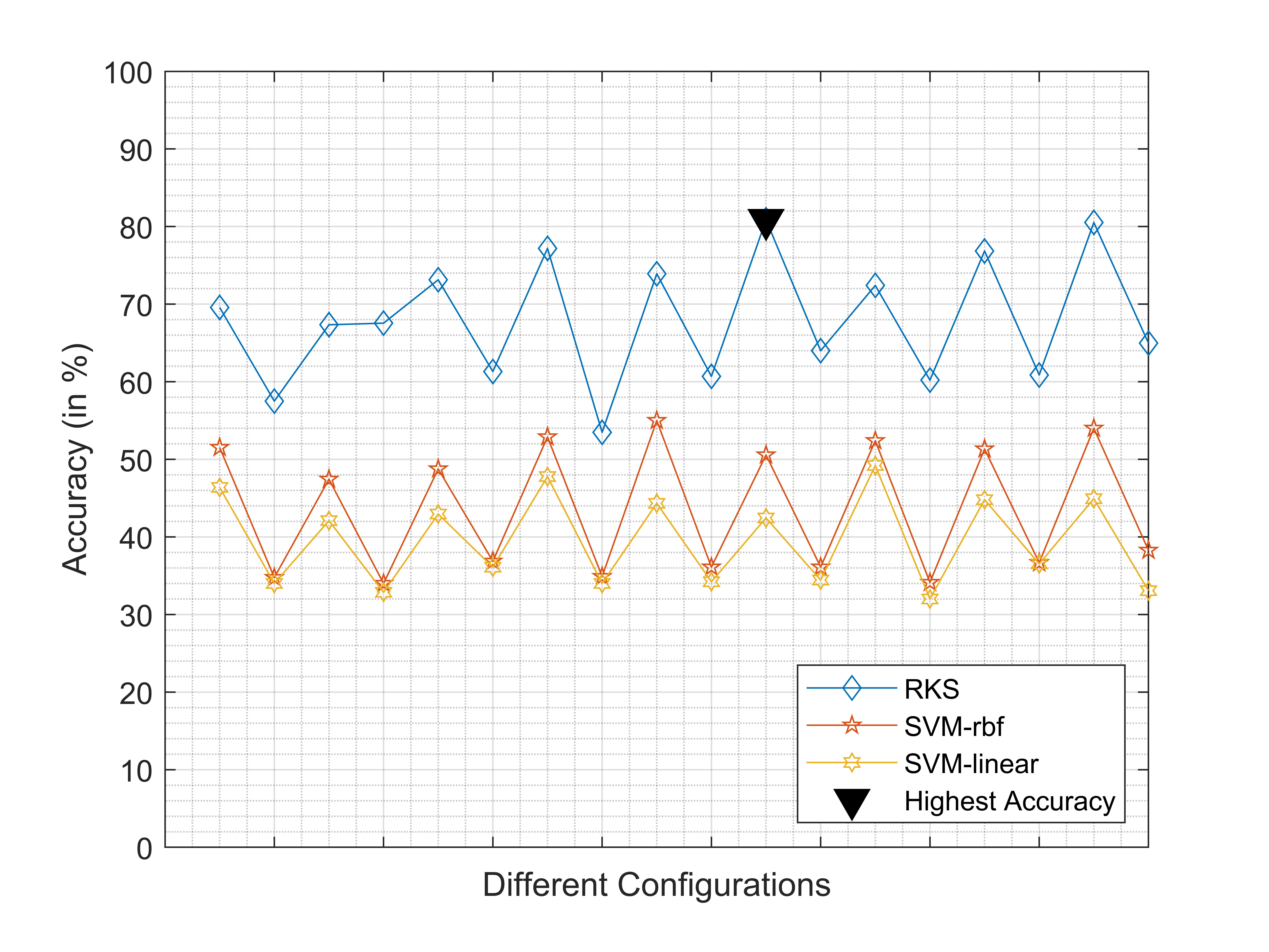}}
\caption{Accuracy plot - RKS Vs SVM-rbf Vs SVM-linear}
\label{fig2}
\end{figure}

The reason why the accuracy tops when the number of Eigenvalues is ``5'' can be explained through Figure \ref{fig3}. After the Eigenvalue index of ``5'', the Eigenvalues cease to change and doesn't contribute much to the underlying dynamics of the image. After the features are extracted, it is given as input for RKS in which it is mapped from 640 to 500. 

\begin{figure}[htbp]
\centerline{\includegraphics[width=90mm,scale=1]{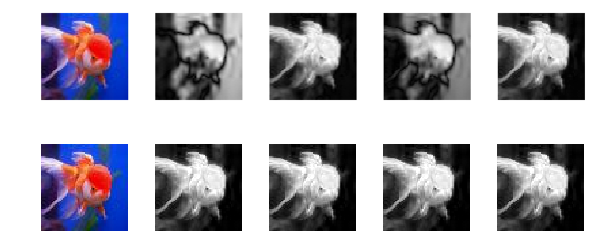}}
\caption{Reconstructed Low-rank ($1^{st}$ row) and Sparse matrix ($2^{nd}$ row) for 5 Eigen Values}
\label{fig4}
\end{figure}

\begin{figure}[htbp]
\centerline{\includegraphics[width=80mm,scale=1]{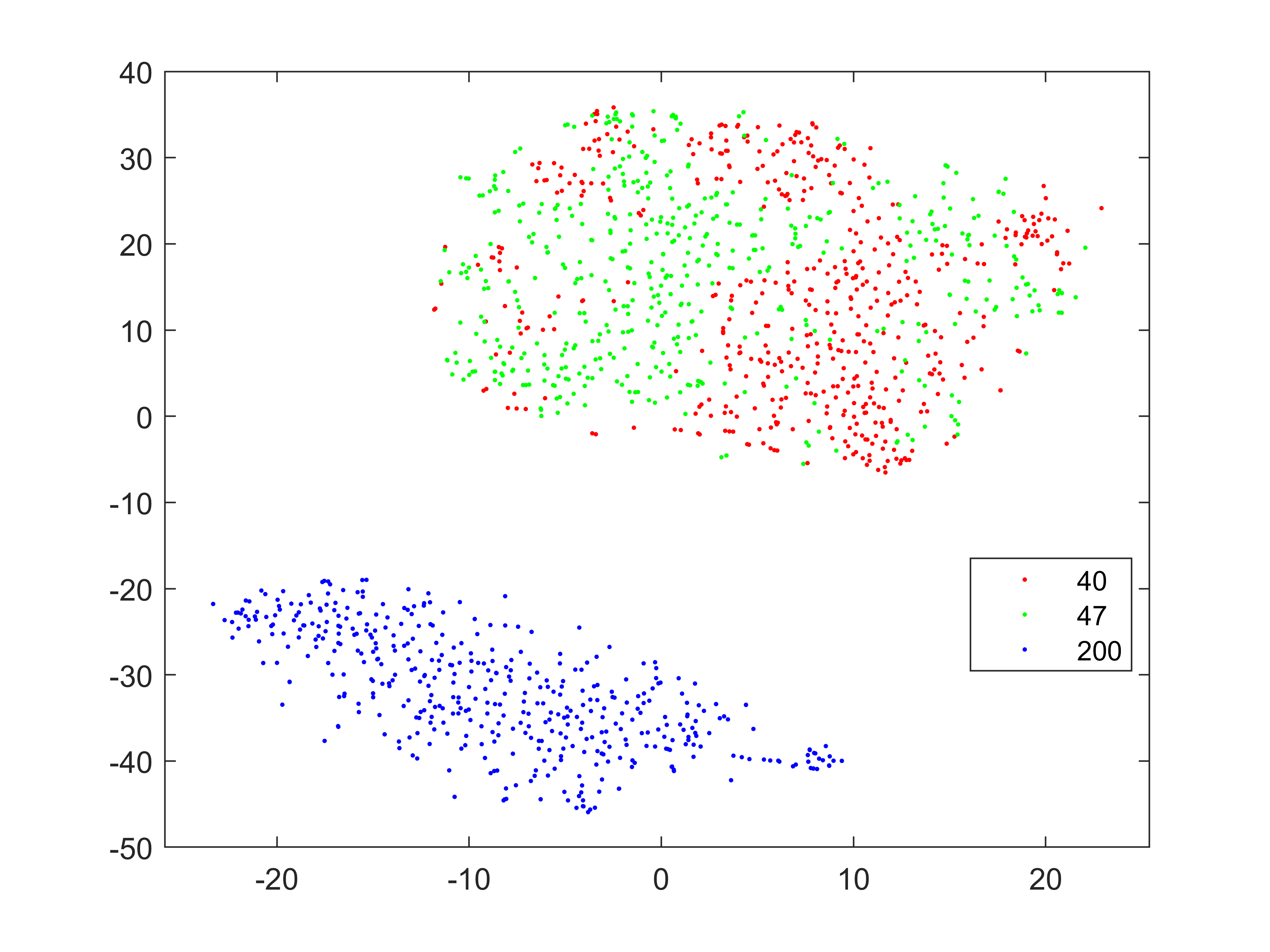}}
\caption{t-SNE plot of 3 classes before applying the proposed approach.}
\label{dmdfig5}
\end{figure}

\begin{figure}[htbp]
\centerline{\includegraphics[width=80mm,scale=1]{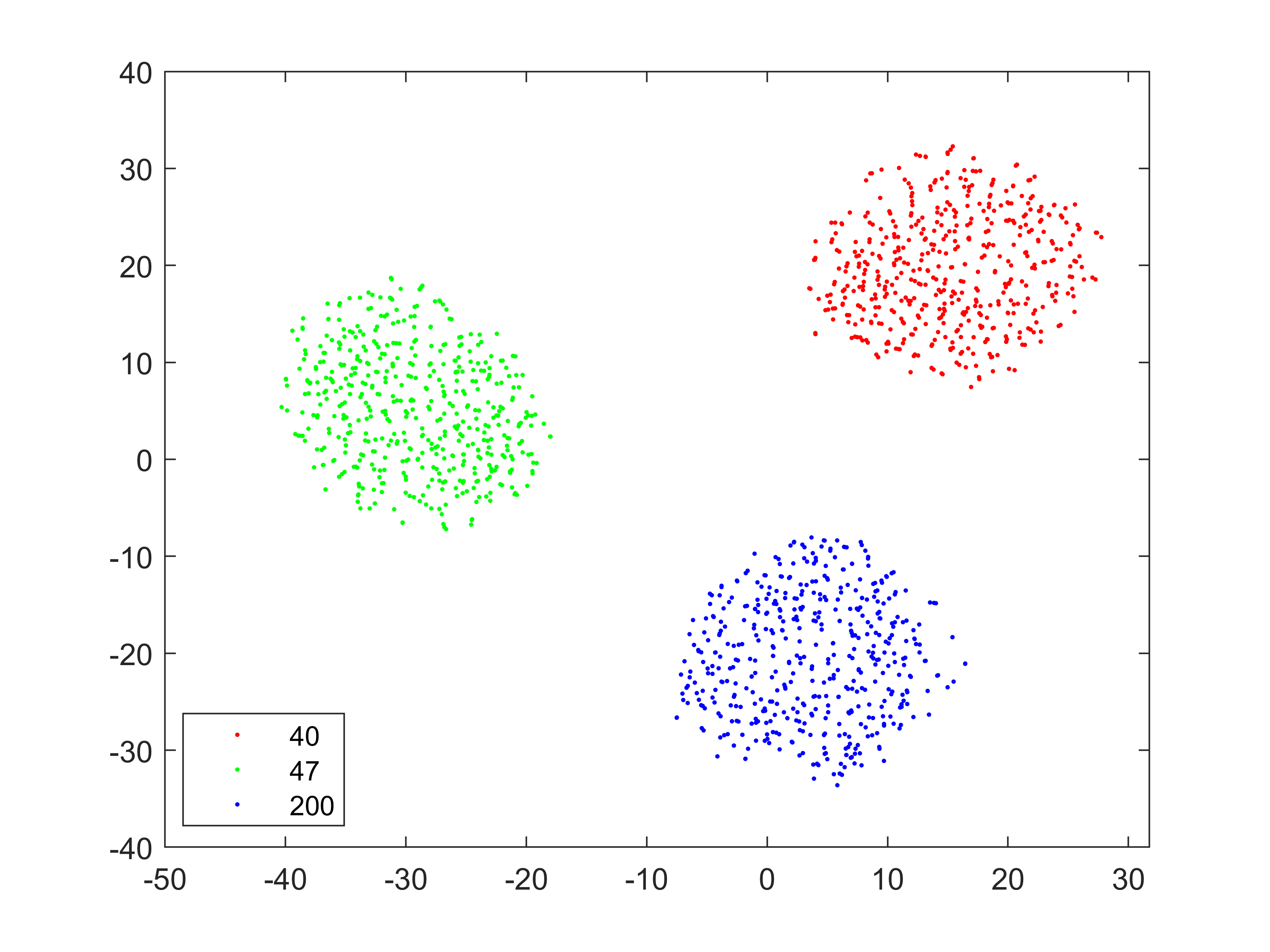}}
\caption{t-SNE plot of 3 classes after applying the proposed approach.}
\label{dmdfig6}
\end{figure}

Figure \ref{fig4} shows the reconstructed image with low-rank and sparse matrix with 5 Eigenvalues which clearly captures the skeleton of the image. This extraction of the skeletal structure is the dynamics captured by applying DMD for feature extraction. t-SNE plot in Figure \ref{2dtsne} provides a better picture of how the extracted DMD features of images arrange themselves distinctive groups flawlessly. Figure \ref{dmdfig5} and Figure \ref{dmdfig6} are the t-SNE plots of 3 classes before and after applying the proposed technique. It is evident from them that, the proposed approach is promising and effective.

\begin{figure}[htbp]
\centerline{\includegraphics[width=80mm,scale=1]{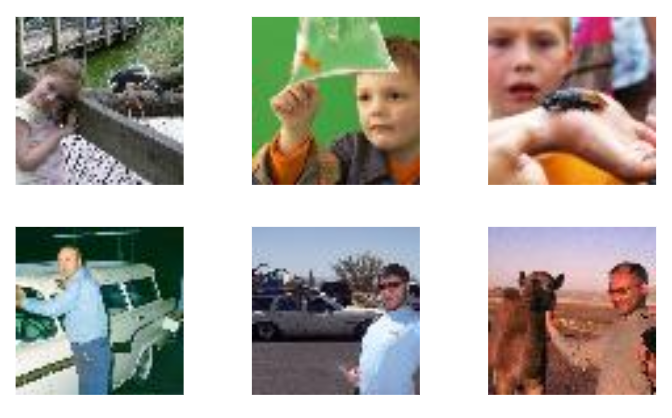}}
\caption{Cases where the proposed approach is ineffective}
\label{fig5}
\end{figure}

The present approach is a novel one and needs more applications-oriented experimental evaluations. There are certain cases like in Figure \ref{fig5}, which has a complex background and where it is difficult to differentiate the foreground of the object of interest from the background, the proposed approach fails to perform. Apart from cases like in Figure \ref{fig5}, the proposed approach is proven to be effective in learning with limited labelled data.

\section{Conclusion}\label{conclusion}
As the world's data generation explodes and due to the reason that manual labelling is highly expensive, it is necessary to develop and explore machine learning architectures that can classify with limited labelled data. The proposed approach in this paper has provided a novel direction in which Dynamic Mode Decomposition based feature can be used in conjunction with a classifier for achieving competing results. As a future scope of this research, the shortcomings of the current proposed architecture can be solved and extended to fast paced applications like Intrusion Detection Systems \cite{51,52} and Data-driven solvers\cite{53} where the data is limited and the classifier must be re-trained frequently on the go.

\end{document}